**Title:** One Print, Many Moves: Monolithic Origami-inspired Folding Actuator for Composable Soft Multi-DoF Systems

**Authors:**
J. Jang[1†], Z. Zhakypov[2†], J. E. Palmer[2], M. Klein[2], J. H. Ryu[1*‡], and A. M. Okamura[2*‡]

**Affiliations:**
[1]IRiS Lab, Department of Civil and Environmental Engineering, Korea Advanced Institute of Science and Technology, Daejeon, Korea.
[2]CHARM Lab, Department of Mechanical Engineering, Stanford University, Stanford, CA, USA.
[†]These authors contributed equally to this work.
*Corresponding authors. Email:jhryu@kaist.ac.kr; aokamura@stanford.edu.
[‡]These authors contributed equally to this work.

**Abstract:** Conventional soft robot actuators excel in compliance, but their uncontrolled deformations compromise accuracy and hinder scaling to multi-degree-of-freedom (DoF) systems. We introduce a MONOlithic ORIGAMI-inspired soft folding actuator design (MONORIGAMI) that establishes a design strategy based on spatially programmed stiffness anisotropy to preserve material compliance along desired folding directions while selectively restricting deformation in unwanted directions**.** The actuator leverages stiffness tiers based on material thickness, patterned in an origami-inspired geometry with facets and creases, converting unconstrained soft deformation into accurate, repeatable, and composable folding motions without additional reinforcements. The design is fully 3D-printable through a single-material, single-print process that requires no assembly. Each actuator serves as a scalable motion primitive, and linking and orienting multiple actuators mechanically programs multi-DoF trajectories. Using the same fundamental module, we demonstrate three 3D-printed soft multi-DoF robotic systems spanning distinct application domains: (1) a compact 4-DoF wearable haptic device for high-fidelity cutaneous feedback in virtual reality (VR), (2) a 3-DoF joystick for kinesthetic feedback in teleoperation, and (3) a modular robotic gripper capable of underwater operation with geometry-encoded grasp trajectories. These systems demonstrate the module's capabilities for compact multi-axis integration, controlled physical interaction, and geometry-programmed operation across different environments. Together, these results show that MONORIGAMI provides a general, composable, accessible, reliable, and scalable platform for high-precision soft multi-DoF robotics, addressing long-standing limitations in both soft actuator design and fabrication.

**One-Sentence Summary:** Spatially programmed stiffness transforms a single-print soft folding actuator into a composable building block for multi-DoF robotic systems.

**Main Text:**

## INTRODUCTION

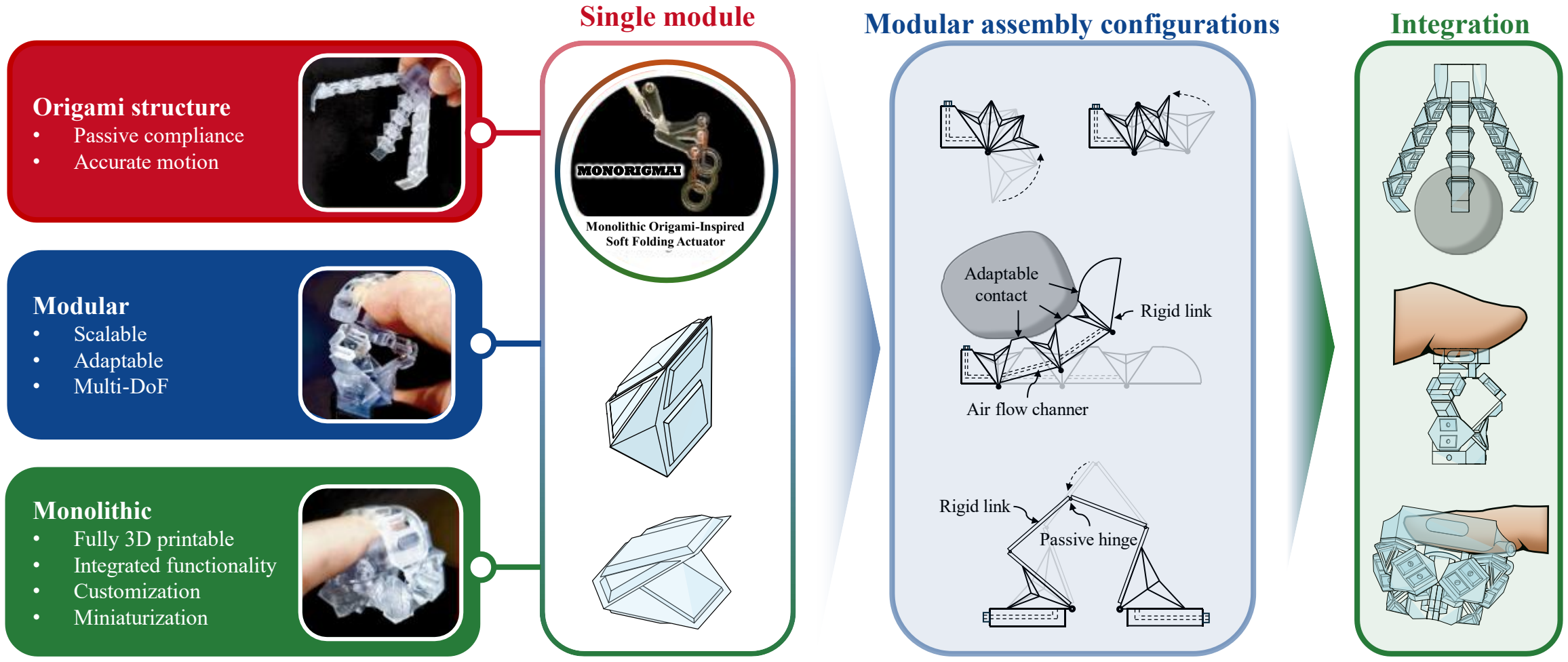

**Fig. 1. The MONORIGAMI Overview.** The MONORIGAMI actuator's origami-inspired geometry spatially programmed stiffness anisotropy selectively suppresses undesired DoF while preserving compliance along prescribed folding directions, enabling accurate and directional motion without added stiffeners. By treating each actuator as a modular building block (single module), multiple units can be linked in serial or parallel to realize predefined actuation trajectories and multi-DoF mechanisms (modular assembly configurations). The entire structure, including the facets, creases, passive joints, and embedded fluidic channels, is printed in a single monolithic step using low-cost 3D printing, enabling extensive customization, miniaturization, and seamless integration across diverse robotic platforms (integration).

Actuator design fundamentally shapes robotic system performance, scalability, and applicability. Conventional rigid robots primarily rely on high-precision rigid actuators, such as electromagnetic (EM) motors and solenoids, which deliver accurate motion and high force output *(1-3)*. However, these actuators require additional mechanical components, including mounts and transmission mechanisms, to connect one joint to another. These connecting structures are typically rigid and heavy, increasing the overall weight and mechanical complexity. Furthermore, their reliance on rigid actuation mechanisms limits the robot's ability to adapt to unpredictable environments, reducing their versatility in applications requiring strength, flexibility, and resilience.

To address these limitations, compliant actuators such as Series Elastic Actuators (SEA) and Antagonistic Elastic Actuators (AEA) have been developed *(4-8)*. These actuators incorporate elastic elements to improve shock absorption, energy storage, and safe interactions with humans. However, their compliance is typically confined to joint flexibility, meaning that they do not provide adaptability at the structural level. As a result, while they offer improved safety and adaptability compared to rigid actuators, they remain constrained in applications requiring distributed compliance across the entire robotic system.

Soft actuators have emerged as an alternative that enables continuous deformation and structural adaptability *(9, 10)*. Various soft actuation technologies, such as pneumatic actuators and material-based actuators (e.g., shape-memory alloys *(11,12)*, dielectric elastomers *(13, 14)*, and liquid crystal elastomers *(15, 16)*), provide inherent compliance, allowing robotic structures to adapt to external forces and unstructured environments. Pneumatic actuation is especially attractive because it can generate large strains and deliver remotely generated power through lightweight tubing to remote points of actuation. Despite these diverse capabilities, conventional soft actuators share a fundamental limitation in applications requiring both compliance and accurate motion. This limitation stems from the inability to restrict undesired deformation, which leads to unintended motion and even instability *(17-20)*. These drawbacks hinder their effectiveness in tasks requiring controlled motion and accurate force application. This also limits their scalability to multi-DoF robotic systems, both in serial configurations where errors propagate along the kinematic chain and in parallel configurations where closed-loop constraints couple joints such that individual errors amplify one another or induce internal stresses. Addressing these challenges is essential for unlocking the full potential of soft actuators and expanding their applicability in advanced robotic systems.

To overcome these limitations, we present a novel MONOlithic ORIGAMI-inspired soft folding actuator (MONORIGAMI) that establishes a design strategy based on spatially programmed stiffness anisotropy. Thickness-defined facets and creases preserve compliance along prescribed folding directions while suppressing off-axis deformation under negative pressure. Unlike conventional EM actuators, which require rigid connecting mechanisms, or existing soft actuators, which are limited in their ability to restrict undesired motion, this actuator achieves a balance between compliance and accurate motion. Each MONORIGAMI actuator functions as a composable motion primitive. By linking and orienting multiple modules in serial or parallel configurations, multi-DoF trajectories can be mechanically encoded (Fig. 1). This composability transforms MONORIGAMI from a standalone actuator into a system-level building block, enabling robotic structures that retain compliance and adaptability while achieving controlled and accurate motion.

Previous approaches to balance compliance and accurate motion rely on a variety of techniques, including self-folding actuation *(21)*, flexible hinge mechanisms *(22)*, and origami-inspired designs *(23-25)*. These approaches combine the inherent structural compliance of soft actuators with relatively stiff components to restrict undesired DoF, thereby improving motion accuracy without sacrificing compliance. However, many rely on separately fabricated hinges or reinforcements, multiple materials, or multistep assembly. More importantly, they have generally focused on individual actuators or mechanisms rather than a composable architecture in which local deformation rules scale predictably to serial and parallel multi-DoF systems. As additional joints or modules are incorporated, hinge stiffness must be finely tuned to maintain directional actuation, and variations in local stiffness, fabrication, or pressure loading may accumulate into large motion errors along serial chains or generate coupled errors and internal stresses in parallel configurations. Thus, a manufacturing strategy that directly encodes deformation constraints into composable soft actuation modules remains lacking.

MONORIGAMI addresses this gap by combining origami structural logic with monolithic additive manufacturing. Locally varied wall thickness creates discrete stiffness tiers across facets and creases, encoding directional compliance and deformation constraints directly into a single material. Additive manufacturing enables complex folding geometries, fluidic channels, and structural constraints to be integrated into a single print without separate reinforcements or post-fabrication assembly (9, 26-29). Unlike prior fully 3D-printed soft robots that focused on digital logic (30), fluidic circuits (31), or geometry-based actuation for dexterous tasks (32), MONORIGAMI establishes thickness-programmed origami stiffness as a general strategy for directional and composable soft actuation. This integration results in a reliable, scalable platform suitable for constructing complex soft multi-DoF robotic systems.

This work makes three primary contributions to the field of soft robotics. (1) A general design principle for directionally constrained soft actuation: We establish spatially programmed stiffness anisotropy as a strategy for suppressing undesired DoF while preserving compliant folding along prescribed directions. This represents a conceptual shift from soft actuators that deform relatively freely to soft actuators whose deformation is designed for accurate, directional motion, reducing the traditional trade-off between compliance and precision. (2) A composable architecture for mechanically programmed multi-DoF motion: Each actuator functions as a composable motion primitive that can be linked and oriented in serial or parallel configurations to produce mechanically programmed multi-DoF trajectories. This transforms the actuator from a component-level device into a platform for constructing complex soft robotic mechanisms, enabling system-level scalability without sacrificing local compliance. (3) A monolithic manufacturing strategy that encodes deformation logic directly into geometry: Leveraging monolithic additive manufacturing, the actuator integrates structural stiffness, fluidic channels, and mechanical constraints through a single-material, single-print process requiring no actuator-level assembly. This fabrication approach provides a reliable and generalizable pathway for producing mechanically programmed soft actuation modules at scale. To demonstrate the breadth of this platform, we validate MONORIGAMI actuators across three robotic systems constructed using fully 3D-printed MONORIGAMI modules (Fig. 1): a 4-DoF wearable haptic device (“FingerPrint”), a 3-DoF delta-style kinesthetic-feedback joystick, and a modular soft gripper with mechanically programmed grasping trajectories capable of underwater operation. These systems respectively demonstrate compact multi-axis integration, controlled physical interaction, and geometry-programmed operation across different environments.

# RESULTS

## Design principle and composable architecture of MONORIGAMI actuators

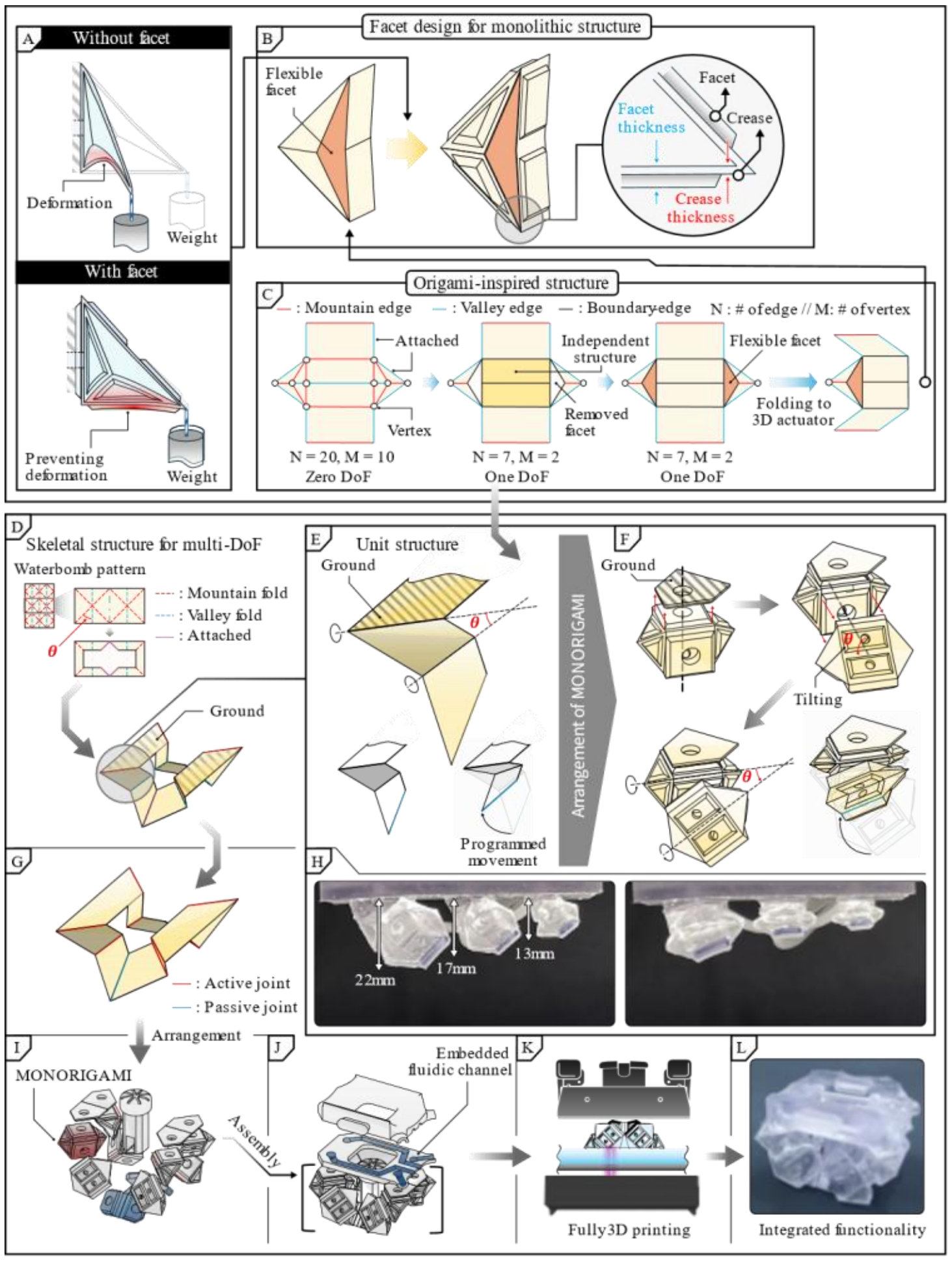

**Fig. 2. Spatial stiffness programming, composable multi-DoF architecture, and monolithic fabrication of MONORIGAMI.** (**A**) Load-bearing capability of origami-inspired structures: Simple chamber-based designs deform under load (top), whereas thickened origami facets provide localized rigidity and improved load resistance (bottom). **(B)** Stiffness tiers created by compliant creases and thickened facet panels enable directional motion without added reinforcements. **(C)** A modified Miura-based crease-facet topology converts a zero-DoF closed pattern into a sealed single-DoF structure. **(D)** A modified waterbomb-based skeleton is used as the kinematic skeleton for multi-DoF motion. (**E**) The unit structure of the waterbomb pattern consists of one ground and two folding lines. **(F)** An integrated fabrication strategy aligns each MONORIGAMI module to its corresponding folding line. (**G**) Passive-active joint configuration enables 4-DoF motion in the modified waterbomb-based skeleton (details in Fig. 5). **(H)** Parametric scalability of MONORIGAMI and skeleton geometries is enabled by high-resolution SLA printing. **(I)** Integration of unit pairs through passive joints to create an example device: a 4-DoF haptic stimulator. (**J**) Integrated fabrication of four paired unit structures with embedded fluidic channels. **(K)** All structural and actuation components are fabricated simultaneously in a single print, preserving their desired configurations. **(L)** The completed print yields a fully integrated, functional multi-DoF device with sealed chambers and embedded fluidic routing.

MONORIGAMI actuators adopt an origami-inspired structure to address the long-standing challenge of achieving both compliance and accurate motion in robotic systems. Conventional soft actuators deform in multiple directions, making their motion difficult to predict or constrain, whereas rigid actuators achieve precision at the expense of adaptability. Origami mechanics mitigate this trade-off through alternating stiff facets and compliant creases that guide deformation along predefined kinematic paths and suppress unwanted motion. This mechanically programmed, directional flexibility enables compliant actuation without sacrificing motion accuracy. Furthermore, when realized as sealed pneumatic chambers, the folding geometry also integrates actuation and simplifies fabrication by removing the need for additional reinforcement or assembly steps.

Origami-based structures achieve directional flexibility through intentional stiffness tiers between compliant creases and load-bearing facets. Instead of simply allowing unrestricted deformation, the geometry localizes compliance at hinge-like regions while reinforcing adjacent panels, thereby defining the kinematic pathway under pneumatic pressurization. This enables predictable motion and controllable force output through mechanically programmed actuation. By tuning the geometry and thickness of crease and facet regions, the actuation range and dynamic response can be precisely customized for specific applications without increasing structural or fabrication complexity.

Building on these principles, we translate the mechanically programmed origami geometry as a monolithic actuator architecture fabricated in a single print cycle (Fig. 2, A–C). Thickened facet panels paired with compliant creases preserve the intended stiffness tiers, enabling accurate, programmable deformation without added reinforcements (Fig. 2B). The discrete stiffness tiers created by the thick facets and compliant creases produce a spatially programmed stiffness anisotropy: deformation is energetically favored along the crease directions and resisted across the facets and in off-axis directions. Implementing this layout through a single-step 3D printing maintains precise geometric relationships among facets, creases, and internal chambers, ensuring consistent motion and enhanced structural robustness (Fig. 2A). Integrating sealing, actuation chambers, and stiffness-encoding geometry into a single fabrication step also enhances design reliability, repeatability, and morphological freedom.

To realize this monolithic actuator architecture in a fully enclosed form while retaining accurate, single-DoF actuation, the crease and facet topology needs to satisfy two conditions simultaneously: (i) the structure remains mechanically sealed under pressurization, and (ii) the origami geometry preserves directional mobility without over constraining the motion. To meet these requirements, we adopted a modified Miura-origami pattern along the actuator side walls (Fig. 2C). Applying a standard Miura pattern theoretically yields zero-DoF, because the closed loop of creases becomes kinematically over constrained. This behavior follows the formulation in *(33)* as follows:

$$\mathrm{DoF} = N - 3M, \quad (1)$$

where N denotes the number of edges and M the number of vertices. For the closed MONORIGAMI crease network, $N = 20$ and $M = 10$. A direct constraint count therefore gives $N – 3M = -10$, indicating that the closed crease network is overconstrained under the assumptions of the adopted mobility formulation. Accordingly, no unconstrained rigid-folding mode is predicted for this closed configuration.

To restore mobility without compromising the enclosure, the center facet of the three-facet Miura unit was removed, converting its edges into free edges and eliminating the associated vertices. This modification transforms the topology from zero DoF to a single kinematic DoF with seven creases and two vertices ($N = 7$, $M = 2$, resulting in DoF = 1). To maintain sealing, the removed facet was replaced with a flexible panel that preserves the enclosed geometry while allowing the programmed motion. This targeted alteration ensures accurate motion while maintaining structural integrity under pressurization.

To extend single-DoF actuation to system-level multi-DoF motion, individual MONORIGAMI modules are assembled into larger kinematic structures. We adopt a modified waterbomb pattern as the skeleton structure, as its crease geometry inherently provides three rotational DoFs when expanded (Fig. 2D–E). The modified waterbomb skeleton is composed of four pairs of unit structures, each with its own programmed deformation pattern (Fig. 2E). Each MONORIGAMI actuator is aligned with a corresponding folding line of the unit structure: the first module is mounted in the same orientation as the ground-connected folding line, after which the second module is tilted relative to the first by the angle between adjacent folding lines (Fig. 2F). This arrangement allows each pair of unit structures to reproduce the intended actuation trajectory (Fig. 2G).

Because MONORIGAMI modules integrate sealed chambers, stiffness-programmed crease–facet geometry, and load-bearing facet panels in a single print, multi-DoF structures can be scaled without extra reinforcement or assembly. The skeleton and MONORIGAMI design can be parameterized and 3D printed at different scales to suit the target motion (Fig. 2H). In this study, devices were fabricated using a commercially available stereolithography (SLA) 3D printer with standard material (Formlabs Form 3 with Flexible 80A resin) *(32)*, whose high resolution and strong interlayer bonding enable thin-walled, air-tight structures with complex pneumatic chambers; comparable additive manufacturing platforms may also be used.

By linking four pairs of unit structures with passive joints (Fig. 2, I and J), we construct a fully functional 4-DoF device in which each joint is independently controlled through its own fluidic channel. Once the geometry is defined, the entire system, including actuators, chambers, and fluidic routing, is fabricated in a single printing step (Fig. 2, K and L), demonstrating that MONORIGAMI enables modular, scalable, fully monolithic multi-DoF robotic architectures.

### Programming actuator performance through crease-facet geometry

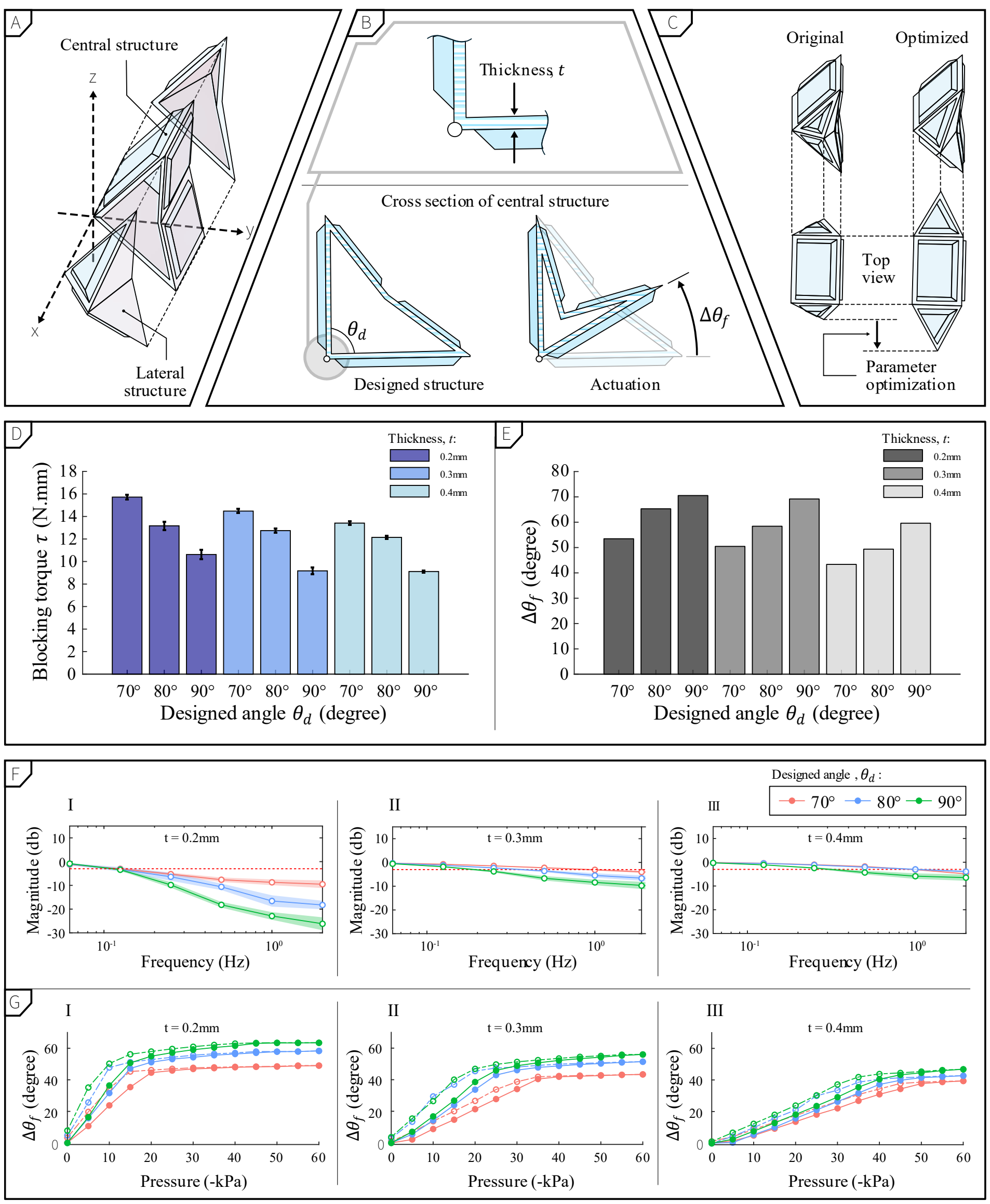

**Fig. 3. Geometry–performance relationships for programming MONORIGAMI actuation.** (A) Geometrical decomposition of the MONORIGAMI into central and lateral structures for defining internal volume and crease parameters. (B) Derivation of the designed folding angle $\theta_d$ from the central geometry. (C) Optimization of lateral-structure parameters to maximize pneumatic energy conversion into mechanical work by minimizing elastic-energy accumulation at the creases. (D) Blocking torque measurements across actuator variants with different crease thicknesses and designed angles. Bars and error bars represent the mean and standard deviation, respectively, over 10 repeated measurements using one actuator specimen for each design. (E) Range of motion (RoM) for all actuator variants under -80 kPa input. (F) Bandwidth characterization across crease thicknesses. (G) Hysteresis evaluation during cyclic pressurization and release.

To quantitatively evaluate the actuation performance of a MONORIGAMI unit, we developed a geometrical model that provides a design-level approximation of how geometry and crease stiffness affect the mechanical energy available for folding. Unlike conventional silicone-based soft actuators, whose energy storage capacity for mechanical work conversion is primarily governed by material stretchability, the MONORIGAMI actuator exhibits minimal extensibility. As a result, its stored energy is primarily determined by the internal volume defined by structural geometry rather than material elongation. At the same time, because compliant crease regions still store elastic energy, actuation performance depends on the balance between pressure-driven energy input and elastic energy at the creases. To guide geometric design, we define a quasi-static design metric representing the pressure work available for folding after subtracting the elastic energy stored in the crease regions. The pressure work is approximated as the product of the applied pressure magnitude and the change in internal chamber volume between the initial and folded configurations. Accordingly, the effective mechanical energy available for folding, $E_{MONO}$, is expressed as:

$$E_{MONO} = E_P - E_{EU} = |\Delta P|\left|V_f - V_i\right| - \frac{1}{2}\Sigma k_i \lambda_i^2 \quad (2)$$

where $|\Delta P|$ is the magnitude of the applied gauge pressure, $V_i$ is the internal chamber volume in the initial, unactuated configuration, $V_f$ is the internal chamber volume in the prescribed folded configuration, $k_i$ is the stiffness of each crease, and $\lambda_i$ is the folding angle. To define $V$, $k_i$, and $\lambda_i$, the geometry was divided into central and lateral structures (Fig. 3A). The target designed angle $\theta_d$ was derived from the central region (Fig. 3B), while in the lateral region, the geometric parameters were numerically optimized by solving a constrained optimization problem using MATLAB's *fmincon* to minimize $E_{EU}$, thereby maximizing $E_{MONO}$ (Fig. 3C). The resulting parameter values are summarized in Table S1, and full derivations and optimization procedures are provided in the Supplementary Materials.

To experimentally characterize and validate the actuator performance, we next generated a set of actuator designs by systematically varying the key geometric parameters that strongly influence $E_{MONO}$, particularly crease stiffness (controlled by crease thickness) and designed angle. Crease thickness was set to three levels (0.2, 0.3, and 0.4 mm), with 0.2 mm being the minimum reliably achievable thickness by the employed fabrication process. The designed angle, determined as described in the Supplementary Materials, was selected as 70°, 80°, and 90° to cover the representative optimized design space. From this parameter set, nine actuator variants were fabricated for experimental characterization. Each actuator was systematically assessed in terms of blocking torque, range of motion (RoM), bandwidth, and hysteresis to comprehensively characterize performance. All actuators used in the following experiments were fabricated using the Form 3 SLA 3D printer (Formlabs Inc.) and Flexible 80A resin. A detailed overview of the test setup and procedures is provided in Fig. S4.

**Blocking Torque.** To assess how model derived design parameters affect torque generation, we measured the blocking torque of each actuator design. Blocking torque represents the maximum resistive torque under constrained motion and indicates how effectively pneumatic energy is converted into mechanical output, including load-bearing and stored elastic strain energy.

Before conducting experiments, we theoretically predicted the maximum torque assuming negligible elastic energy loss in the material. As shown in Fig. S2, the theoretical model predicted that maximum torque would be proportional to the designed angle $\theta_d$ within the range of 0° to 120°, with larger designed angles yielding higher blocking torque due to greater volume change during actuation. However, experimental results revealed a contrary trend. Fig. 3D and Figs. S5–S7 (A, C, and E) show blocking torque measured across different crease thicknesses and designed angles under square-wave negative pressure inputs (−20, −40, −60, and −80 kPa), while motion was restrained and reaction forces were recorded. Across all thickness conditions, the highest measured blocking torque was 15.72 N·mm for a crease thickness of 0.2 mm and $\theta_d$ = 70°, and the lowest was 9.11 N·mm for a crease thickness of 0.4 mm and $\theta_d$ = 90°, contradicting the theoretical prediction.

This discrepancy arises from the dominant effect of elastic energy loss on blocking torque. Thicker creases produced lower torque at the same pressure due to greater elastic energy storage and reduced pressurizable volume. Although increasing $\theta_d$ raises theoretical volume change, the elastic deformation and associated energy dissipation in the material increase more substantially, leading to reduced blocking torque. This finding highlights the importance of considering material elastic properties alongside geometric parameters when designing soft actuators for force-generation applications.

**Range of Motion.** Geometric parameters affect actuator motion capacity, quantified by its RoM. In multi-DoF systems, RoM plays a critical role, as it directly affects dexterity and workspace and consequently constrains the coordination of multiple modules for achieving desired spatial configurations. Fig. 3E and Figs. S5-S7 (B, D, and F) show the RoM for all variants was measured under a -80 kPa square wave input, with angular displacement tracked using color markers.

The results reveal a clear dependence of RoM on both crease thickness and designed angle. The largest RoM (70.53°) was observed for a crease thickness of 0.2 mm and a designed angle of 90°, and the smallest RoM (43.35°) occurred for a crease thickness of 0.4 mm and a designed angle of 70°. Thicker creases reduce RoM due to higher stiffness, and smaller designed angles limit the maximum folded angle, lowering RoM. Notably, increasing $\theta_d$ increased RoM but decreased the measured blocking torque, demonstrating a design trade-off between motion range and output torque.

**Bandwidth.** Whereas blocking torque and RoM characterize the static and quasi-static performance of the actuator, dynamic responsiveness is captured by bandwidth. Unlike the former metrics, which are primarily influenced by the designed angle $\theta_d$, bandwidth depends more strongly on crease thickness $t$ (Fig. 3F). Bandwidth was measured by applying −80 kPa square-wave pressures at input frequencies from 0.0625 to 2 Hz, with ten repetitions per frequency.

Actuators with $t$ = 0.2 mm fell below −3 dB at frequencies under 0.125 Hz, showing exponential amplitude decay as frequency increased. In contrast, actuators with $t$ = 0.3 mm and $t$ = 0.4 mm maintained responses above −3 dB up to approximately 0.25–1 Hz and exhibited more gradual amplitude decay. These results suggest that thicker creases enhance bandwidth by increasing elastic energy storage, thereby accelerating release phases. However, since bandwidth reflects both response and release timing, it does not fully represent actuation speed. A more detailed performance analysis is provided in the "Load-bearing capacity and actuation speed" section.

**Hysteresis.** To complete the cyclic performance evaluation, we quantified hysteresis in free motion to assess the effects of material dynamics and crease thickness on actuation repeatability (Fig. 3G). Pressure was reduced in −5 kPa increments while tracking the folding angle through pressurization and release. Actuators with $t$ = 0.2 mm fully folded at approximately −20 to −25 kPa, whereas those with $t$ = 0.3 mm required about −35 kPa. For $t$ = 0.4 mm, complete folding occurred at −45 to −50 kPa, reflecting greater elastic energy storage in thicker creases. Although minor hysteresis appeared during release, likely due to viscoelastic material response, its magnitude was small across all designs. This indicates that the MONORIGAMI actuator enables repeatable motion with minimal energy loss under cyclic actuation.

These results establish a geometry–performance relationship for MONORIGAMI modules. Thinner creases reduce elastic resistance and increase blocking torque and RoM, whereas thicker creases improve elastic recovery and dynamic bandwidth. The designed folding angle primarily determines the achievable motion range, while its influence on torque is coupled with elastic-energy accumulation. Thus, crease thickness and designed angle provide complementary geometric variables for programming the force, motion, and dynamic response of each motion primitive.

## Load-bearing capacity and actuation speed

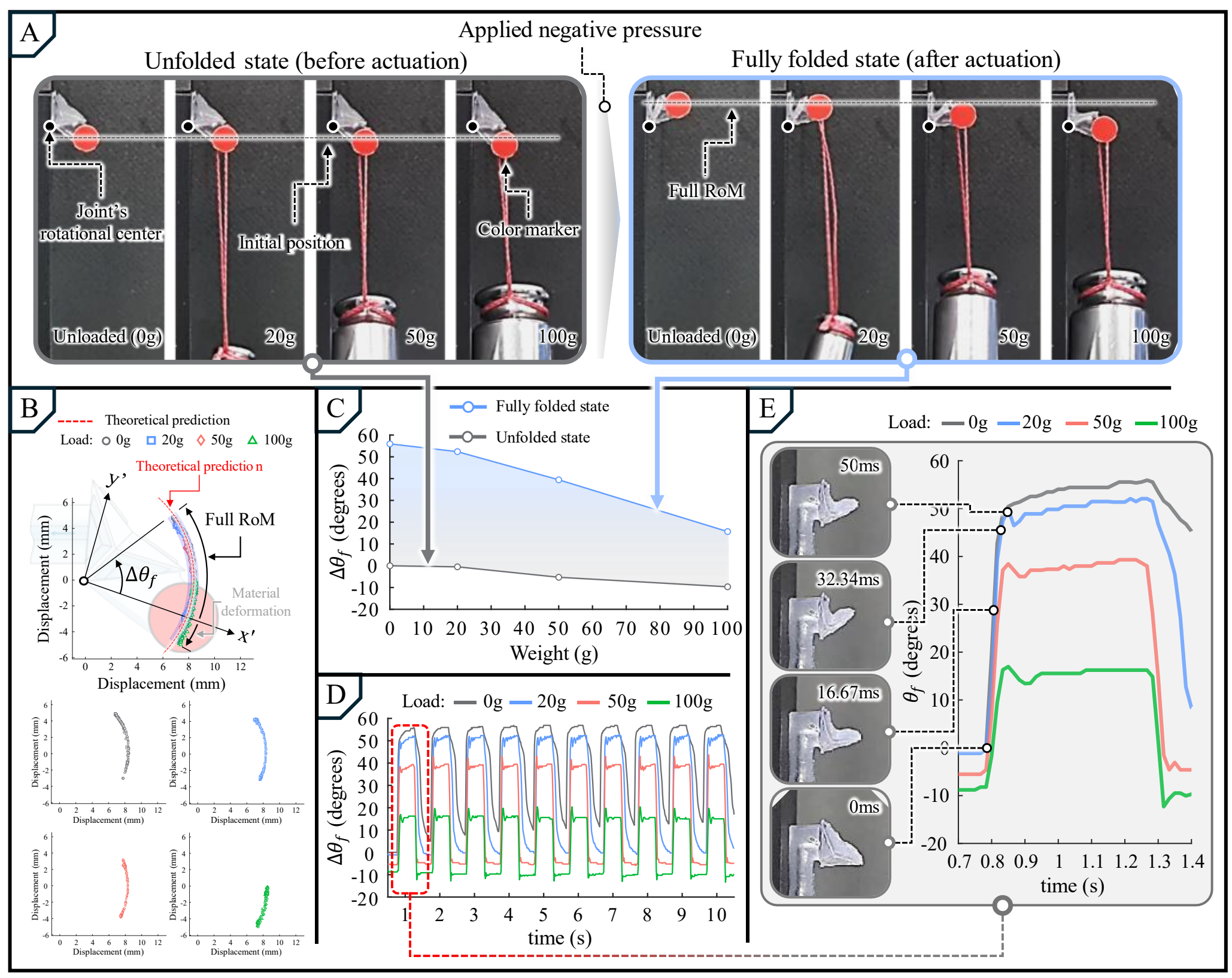

**Fig. 4. Directional motion robustness and actuation performance under external loads. (A)** MONORIGAMI actuator ($t$ = 0.3 mm, $\theta_d$ = 80°) with loads of 0 g, 20 g, 50 g, and 100 g suspended 8.5 mm from the joint's rotational center, shown in unfolded state (before applying negative pressure) and fully folded state (after applying negative pressure). Color markers tracked displacement throughout the experiments. **(B)** Rotational motion accuracy under varying loads. The x' axis defines the reference frame, with negative values indicating material deformation. **(C)** Range of motion versus suspended weight. RoM decreased from 56° (no load) to 26° (100 g load), while material deformation increased to −10° at maximum load. **(D)** Repeatability of actuation speed and RoM under varying constant load conditions. The actuator demonstrates high consistency across multiple trials, confirming constant torque generation throughout the motion. **(E)** Actuation-speed comparison across loading conditions.

The central premise of spatially programmed stiffness anisotropy is that the actuator should preserve its prescribed folding direction even when external forces introduce competing deformation modes. We therefore evaluated the trajectory accuracy, off-axis material deformation, and response speed under increasing external loads. Having characterized MONORIGAMI in free motion, we evaluated the accuracy and response speed under external loads. Load-bearing performance is essential for applications requiring consistent motion under interaction forces, and soft pneumatic actuators (SPAs) often rely on rigid components, increasing fabrication complexity. In contrast, the MONORIGAMI actuator achieves intrinsic load bearing through its origami geometry, which restricts undesired DoF without rigid reinforcements, enabling consistent motion and fast response under load even when operating against external loads.

To evaluate these properties, we measured displacement of the actuation point under different loads. An actuator with $t$ = 0.3 mm and $\theta_d$ = 80° was driven by -80 kPa square wave pressure at 1 Hz. Loads of 0 g, 20 g, 50 g, and 100 g were suspended 8.5 mm from the joint's rotational center, and deformation and response time were tracked using color markers. Each condition was repeated ten times under identical settings.

Material deformation was defined by establishing an x' axis from the joint's rotational center to the color marker under no load, with positions below zero indicating material deformation. As load increased, full RoM decreased exponentially due to reduced internal volume at larger folding angles and lower generated torque (Fig. 4B). Material deformation increased from 0° with no load to -10° at 100 g, reducing RoM from 0° to 56° to approximately -10° to 16°.

Despite a slight increase in material deformation with added weight, the MONORIGAMI actuator maintained accurate rotational motion even when operating under external loads. Across ten trials, the endpoint displacement error remained within ±5% of the theoretical trajectory when normalized by the unloaded full-scale displacement (Fig. 4B). The actuator also showed high repeatability in actuation speed and RoM (Fig. 4D), indicating consistent torque generation under constant load.

As discussed in the section on "Actuator geometry optimization and experimental performance characterization", all MONORIGAMI designs exhibited a -3 dB bandwidth below 1 Hz, suggesting slow actuation performance. However, the actuator reached full folding in about 50 ms under all load conditions (Fig. 4E), indicating that the low bandwidth was primarily due to release time rather than response time. This implies that bandwidth could be improved through actuator configuration in real-world applications under load, as further demonstrated in the following section on "a 4 DoF wearable haptic device enabled by MONORIGAMI actuation".

In conclusion, these load-bearing experiments demonstrate that the MONORIGAMI achieves a rare combination of intrinsic motion accuracy, high repeatability, and fast actuation, even under substantial external loading, thereby validating its suitability as a robust building block for scalable soft multi-DoF systems. These results confirm that origami inspired mechanical programming stabilizes directional motion and mitigates undesired deformation, enabling reliable performance under disturbance.

**4-DoF wearable haptics enabled by lightweight, compliant, and high-accuracy MONORIGAMI actuation**

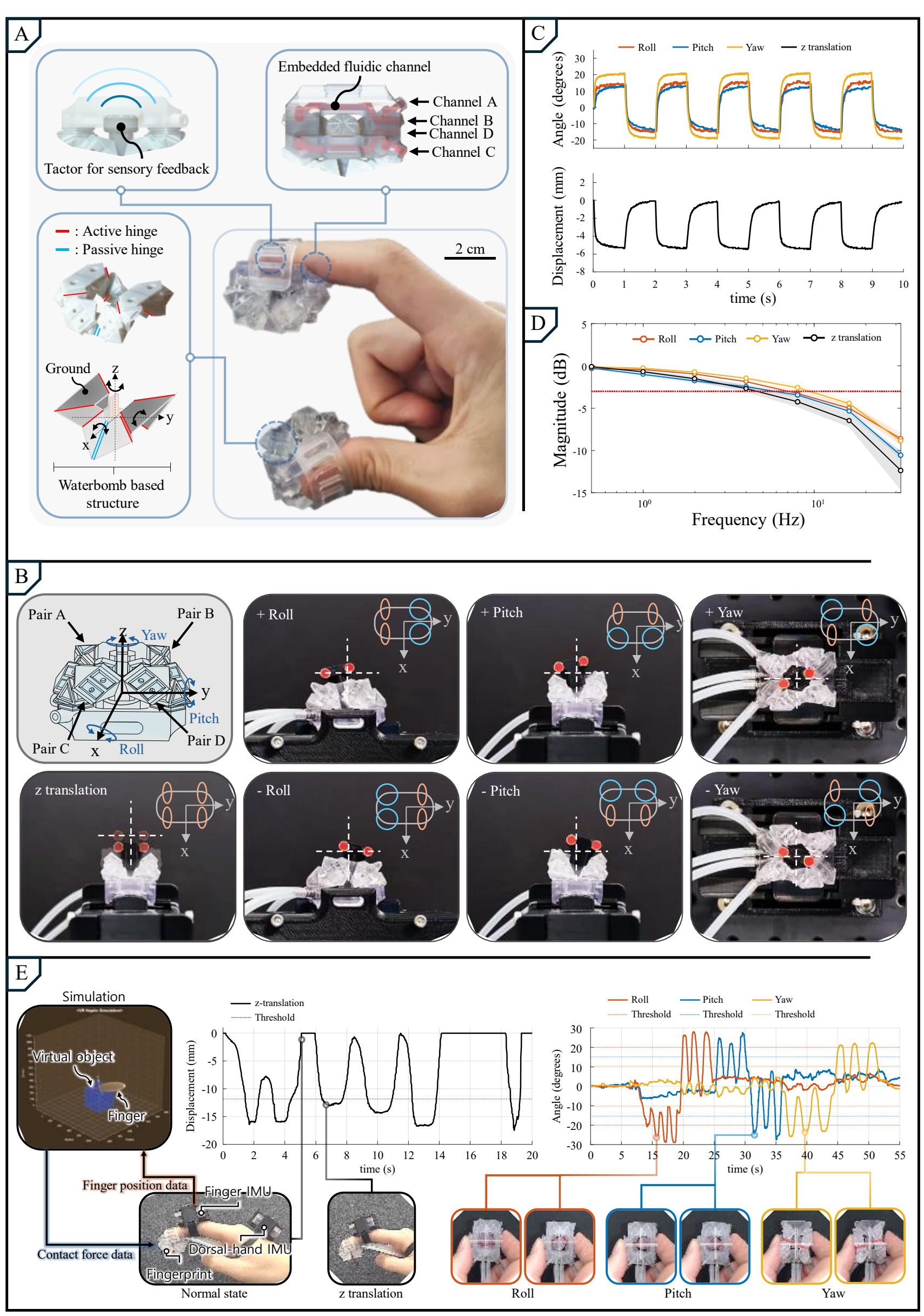

**Fig. 5. Orientation-programmed serial–parallel integration of MONORIGAMI modules for compact 4-DoF wearable haptic device (FingerPrint) for delivering cutaneous feedback in virtual reality (VR) environments. (A)** modified waterbomb-based skeleton integrating four pairs of serially connected MONORIGAMI actuators, with passive hinges, embedded fluidic channels (A–D), and a tactor fabricated monolithically for compact multi-DoF cutaneous actuation. **(B)** Inherent compliance of the MONORIGAMI modules enables an additional z-axis translational DoF when all actuator pairs are activated simultaneously. **(C)** Measured workspace of the device, achieving ±17° roll and pitch, ±20° yaw, and 4.7 mm z-axis translation. **(D)** Bandwidth characterization showing >50% actuation magnitude above 4 Hz across rotational DoFs. **(E)** Integration of the FingerPrint with a VR control framework, where IMU-based fingertip pose and contact forces are mapped to actuator commands for real-time cutaneous feedback.

Having established MONORIGAMI actuator's lightweight, compliant, and accurate actuation performance, we further demonstrated its applicability by developing a compact 4 DoF device called *FingerPrint* for skin deformation feedback in virtual reality (VR). Compared with our earlier prototype *(34)*, the updated system provides a larger end effector workspace with similar yaw RoM and more balanced roll and pitch RoM, while reducing the overall size to two thirds of the previous version.

Among the diverse VR haptic approaches, skin deformation devices deliver localized cutaneous feedback through normal indentation and lateral or tangential stretch. By stimulating the fingertip, a region rich in mechanoreceptors, they enable compact wearable systems that provide immersive and precise tactile sensations.

Despite their potential, traditional skin deformation haptic devices with rigid actuators are often bulky and insufficiently compliant, while soft actuators typically lack precise stiffness control. MONORIGAMI actuators address these limitations by offering directional compliance, programmable motion, and high bandwidth actuation, all in a monolithically printed form that reduces weight and complexity.

The FingerPrint device integrates four serial pairs of MONORIGAMI actuators in a modified waterbomb-based skeleton (Fig. 5A). Each actuator serves both as an actuator and structural joint, allowing the system to achieve roll, pitch, and yaw motions via selective actuator pair activation. Due to inherent compliance and modular stiffness, simultaneous activation of all pairs produces a fourth DoF in z-axis translation. The system fits within $40 \times 40 \times 40$ mm$^3$ and weighs 5 g. Fluidic channels and the tactor were directly integrated into the monolithic structure fabricated using single-material 3D printing. This eliminates the need for connectors and assembly, reducing mechanical failure points and enabling seamless device miniaturization.

A key feature is that the inherent compliance of the MONORIGAMI modules enables additional z-axis translation when all actuator pairs are activated simultaneously. Unlike a rigid modified waterbomb-based skeleton, the compliant modules allow elastic deformation, enabling the FingerPrint to provide roll, pitch, yaw, and z translation using four independent fluidic channels (Channels A-D).

Performance testing showed that the FingerPrint achieves ±17° in roll and pitch, ±20° in yaw, and 4.7 mm in z-axis translation (Fig. 5C). These values correspond to the typical orientation changes of a fingertip during object manipulation and surface exploration, indicating that the device can reproduce naturalistic cutaneous motion patterns. Rotational DoFs retained over 50% actuation magnitude above 4 Hz (Fig. 5D), exceeding the response rate required for high-speed touch interactions in VR (e.g., tapping, sliding, button pressing). This demonstrates that the device can deliver fast and crisp tactile cues rather than sluggish or dampened motion. By integrating structure and actuation in a compact monolithic design, the MONORIGAMI allows haptic devices like FingerPrint to meet the competing demands of speed, precision, and wearability, demonstrating its unique suitability for immersive, multi-DoF VR interaction.

To assess feasibility for VR, we integrated FingerPrint with a virtual simulation that generated tactile feedback during object interaction. Finger pose was captured using IMUs on the finger and dorsal hand. Roll, pitch, and yaw were mapped one-to-one to the simulated finger, while 1 degree of dorsal hand motion produced 10 mm of z-axis translation. Feedback was triggered only upon virtual contact, and the resulting pose data were sent to an MCU that actuated selected MONORIGAMI pairs when predefined thresholds were exceeded. Thresholds were -12 mm in z, ±20° in roll, ±15° in pitch, and ±10° in yaw. Corresponding motions were generated to provide directional cutaneous feedback. The framework was further demonstrated in a real VR environment, with details provided in the supplementary materials. As illustrated in Fig. 5E, z-axis translation occurred in the FingerPrint when the z-direction threshold was exceeded. Similarly, when the roll, pitch, and yaw data exceeded their respective thresholds, corresponding movements were generated by the FingerPrint device to provide directional cutaneous feedback to the user.

**A Compact 3-DoF Kinesthetic-Feedback Joystick for Teleoperation Enabled by Accurate and Compliant MONORIGAMI Actuation**

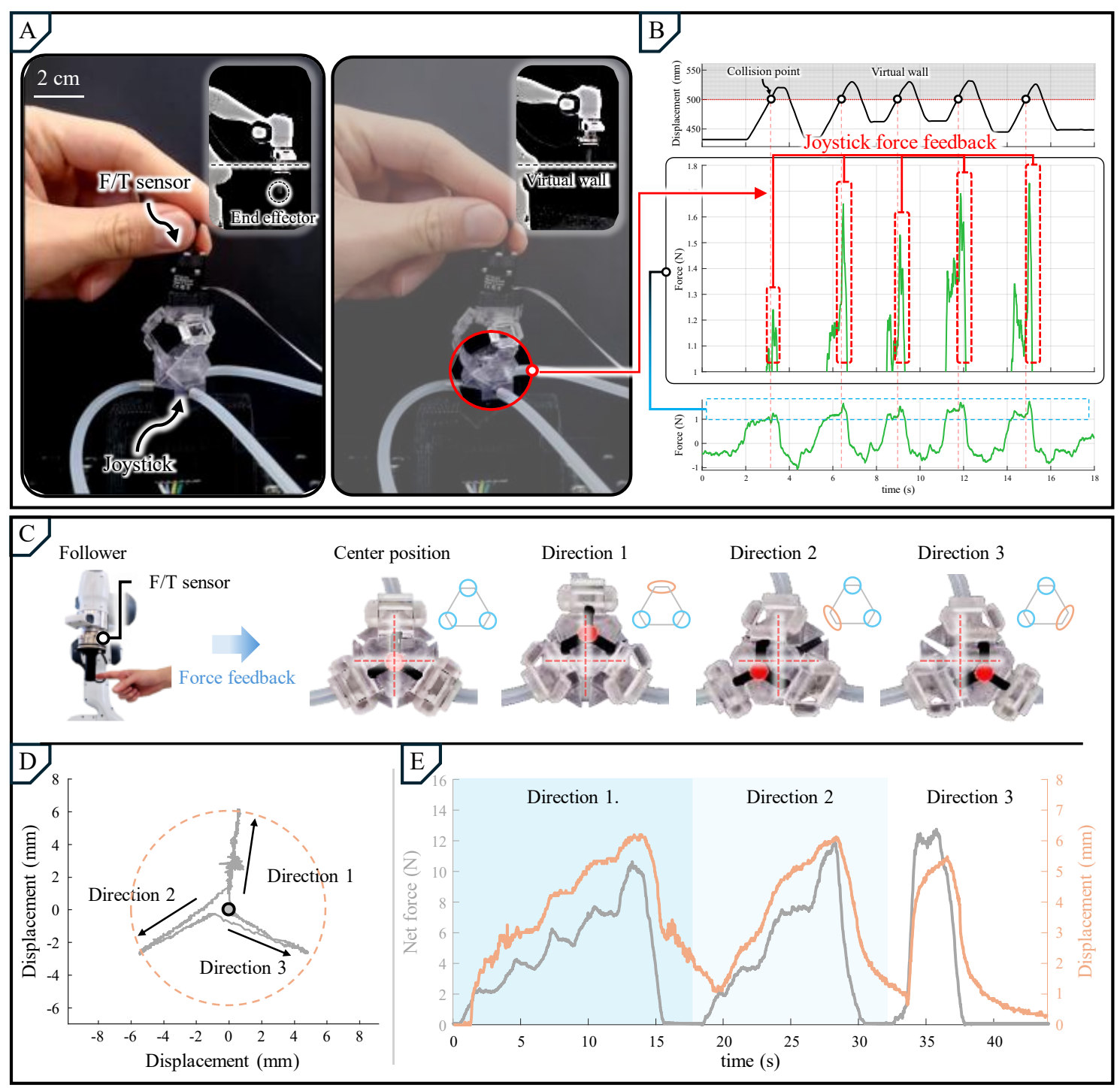

**Fig. 6. Parallel integration of MONORIGAMI modules into a 3-DoF soft joystick for high-fidelity kinesthetic teleoperation with directionally resolved kinesthetic feedback. (A)** Experimental setup for virtual wall evaluation, where contact with predefined workspace boundaries triggers immediate kinesthetic feedback to the operator. **(B)** Real-time kinesthetic-feedback responses during repeated virtual wall encounters, showing instantaneous kinesthetic cues (0.2–0.55 N). **(C)** Environmental contact experiment in which forces measured at the follower side are reproduced by the joystick, enabling vector-accurate kinesthetic feedback along three axes. **(D)** Joystick displacement in three Cartesian directions during the environmental contact test, showing accurate directional response to externally applied forces. **(E)** Comparison between joystick displacement and the net force measured at the follower, demonstrating that the joystick precisely tracks the applied force magnitude and direction across all tested axes.

Kinesthetic-feedback joysticks (leader) are vital in teleoperation for transmitting remote interactions to the operator, yet delivering high fidelity multi-DoF feedback in a compact compliant form remains challenging. Rigid-actuated joysticks are precise but bulky and complex, while soft actuators often lack directional stiffness for accurate force transmission. The MONORIGAMI actuator addresses these limitations through modular programmable stiffness and monolithic fabrication, enabling lightweight compact interfaces with accurate multi-DoF motion for human interaction. Its directional compliance and high force to weight ratio further support sensitivity and robustness in dynamic teleoperation.

To demonstrate these capabilities, we developed a 3-DoF soft kinesthetic feedback joystick using three MONORIGAMI actuators in a delta-inspired parallel configuration *(37, 38)*. Each actuator serves as both structure and actuation element, and the entire device, including frame and pneumatic channels, is fabricated in a single 3D printing process with Flexible 80A material. This monolithic design removes assembly steps and enables a compact 35 × 35 × 35 $mm^3$ device weighing about 3 g.

The joystick generates kinesthetic feedback by selectively activating the MONORIGAMI actuators in response to environmental forces sensed at the follower side (Fig. S9). Conventional rigid joysticks often transmit forces with noticeable latency due to inertia. In contrast, the MONORIGAMI's low-mass actuation and high force-to-weight ratio allow force cues to be produced rapidly. As a result, contact events, resistance, and directional cues are conveyed with high clarity during telemanipulation.

To assess kinesthetic-feedback performance, we performed a virtual wall test and an environmental contact test. The virtual wall technique is commonly employed in teleoperation to improve task-level accuracy and repeatability while reducing operator fatigue. By providing immediate kinesthetic feedback upon boundary contact, it prevents the follower from entering kinematic singularity-prone regions or colliding with the surrounding environment. In this experiment, the follower operated within a bounded workspace, and real time kinesthetic feedback was delivered when it approached or contacted a boundary (Fig. 6A). This prevented entry into singular regions or unintended collisions. During five consecutive contacts, upon contact detection at the follower side, the joystick produced kinesthetic feedback of 0.2 to 0.55 N promptly after contact detection, allowing users to immediately recognize clear boundary contacts (Fig. 6B). The immediate feedback allowed operators to quickly adjust trajectories and avoid constraint regions, confirming reliable collision detection performance.

In the environmental contact experiment, a 6 DoF force/torque (F/T) sensor on the follower measured interaction forces during contact, which were transmitted in real-time to the joystick. The joystick reproduced corresponding kinesthetic feedback by actuating selected MONORIGAMI actuators (Fig. 6C). To evaluate force transmission fidelity, contact forces were applied using stepwise increases (directions 1 and 2) and square-wave patterns (direction 3). The joystick displacement closely matched the applied force profiles in both temporal and amplitude dimensions (Fig. 6D, E). Notably, the joystick reproduced directional reversals of follower-side forces in synchronization across all three axes, indicating that haptic feedback preserves and transmits vectorial information rather than merely scalar force magnitudes. This vector-level correspondence under dynamic loading conditions allows operators to directly perceive subtle contact dynamics such as stiffness, friction, slope variations, and impact transients through the joystick, enabling precise manipulation in teleoperation.

The directional stiffness of each MONORIGAMI can be tuned through geometry and material thickness, allowing task-specific feedback customization. Its compliant behavior enables passive return to neutral without external springs, simplifying the design. By providing a compact, multi-DoF kinesthetic-feedback system with accurate response, safe operation, and simple fabrication, the MONORIGAMI actuator advances the usability and integration of small, lightweight robotic joysticks with tunable compliance in next-generation teleoperation systems.

**Modular soft robotic gripper with mechanically programmed finger actuation**

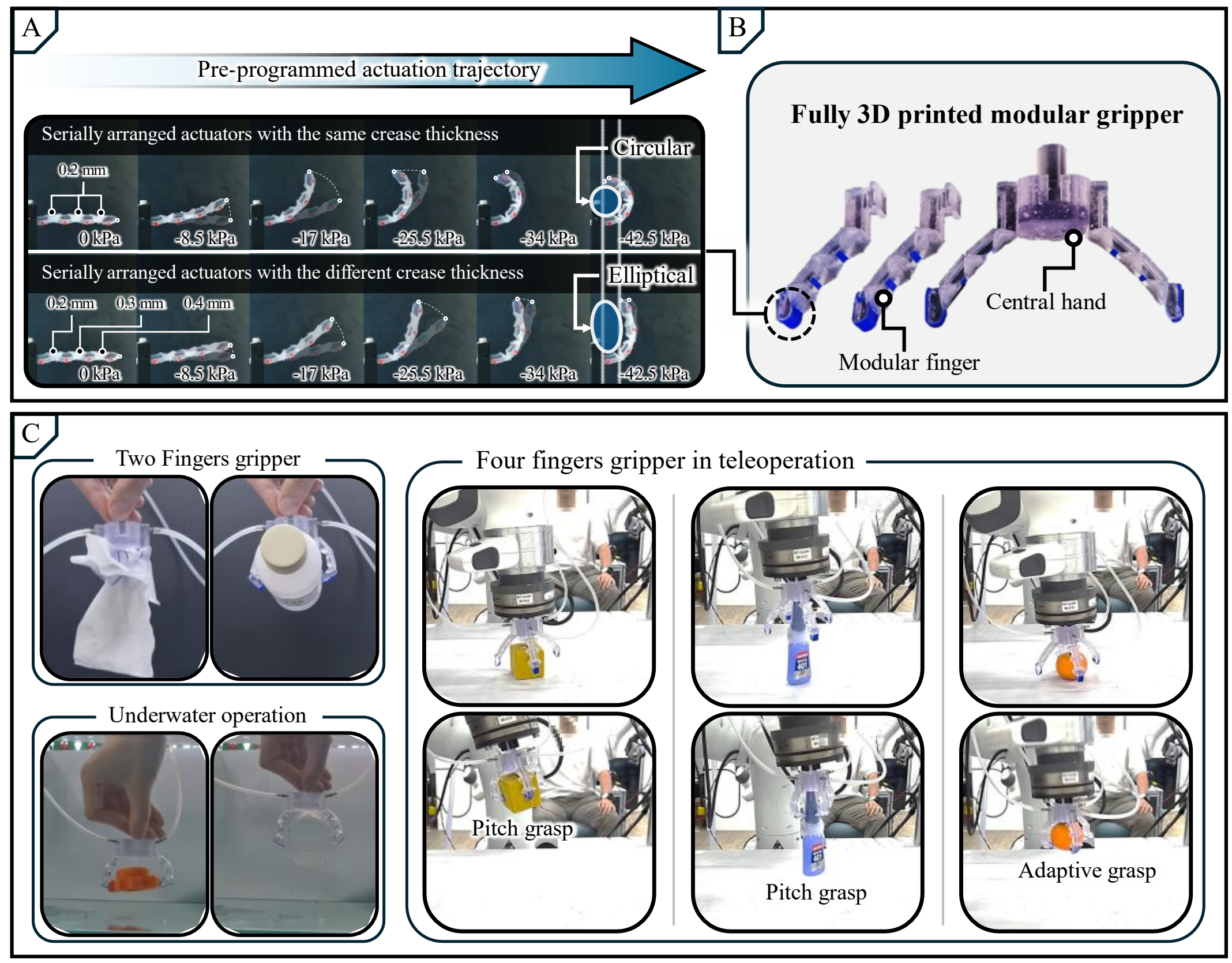

**Fig. 7. Heterogeneous serial integration of MONORIGAMI modules for a soft gripper with tunable, geometry-programmed grasp trajectories. (A)** Programmable finger trajectories generated by connecting MONORIGAMI actuators with different crease thicknesses $t$ and designed angles $\theta_d$ in series. **(B)** Modular multi-finger gripper design in which each finger's motion profile is determined by its actuator configuration, allowing finger modules to be freely replaced for task-specific grasping strategies. **(C)** Demonstration of the gripper's versatility, including underwater operation enabled by its monolithic, electronics-free design, as well as task execution using both two-finger and four-finger configurations.

The tunable nature of MONORIGAMI actuators, achieved through variations in crease thickness $t$ and design angle $\theta_d$, enables the serial connection of actuators with distinct characteristics to generate mechanically programmed trajectories (Fig. 7A). In modular grippers, this tunability and scalability enable rapid fabrication and replacement of task specific fingers, expanding operational capability. Unlike conventional soft grippers that require additional rigid components for modularization due to their excessive compliance *(41)*, each finger can be fabricated as a monolithic 3D-printed module without internal assembly, and the modules can subsequently be attached to a common hand base. Furthermore, the absence of electronic components enables the gripper to operate in underwater environments without specialized waterproofing treatments.

The modular gripper consists of four modular fingers and a central hand (Fig. 7B). The grasping motion of each finger depends on the design of the actuators. Fingers equipped with identically designed actuators exhibited uniform bending at each joint, resulting in a circular grasping motion under the same pressure conditions. In contrast, fingers with differently designed actuators showed varying joint angles under identical pressure, producing an elliptical grasping motion (Fig. 7A). This flexibility allows the contact points of the fingers to be preconfigured according to the object being grasped. By designing actuators with different characteristics, the gripper's performance can be enhanced through modularity, enabling the fingers to be replaced freely to suit different tasks.

Experimental demonstrations validate the gripper's adaptability through efficient task execution by reconfiguring between two-finger and four-finger configurations according to object geometry (Fig. 7C). The soft-material-based construction and pneumatic actuation enable seamless operation in submerged conditions without waterproofing treatments. These results demonstrate the proposed actuator's high versatility and its potential to enhance the adaptability and performance of soft robotic systems across diverse application domains.

## DISCUSSION

MONORIGAMI differs from many conventional soft actuators in how softness is used. Rather than reducing compliance through uniform reinforcement, the proposed design organizes compliance spatially through a thickness-defined stiffness contrast between compliant creases and load-bearing facets. This distribution preserves deformation along prescribed folding directions while suppressing off-axis motion. Softness therefore becomes a geometrically programmed property that defines actuator behavior. Under external loading, the actuators continued to follow their prescribed rotational paths, although increasing load reduced the achievable range of motion and increased material deformation. These results indicate that MONORIGAMI does not eliminate the intrinsic compliance of the material, but instead redistributes it into mechanically defined motion pathways.

The directionally constrained motion of each MONORIGAMI module enables system-level composition. By varying module orientation, serial–parallel connection topology, crease thickness, and designed folding angle, modules can be combined to generate multi-axis motions and nonuniform trajectories, establishing MONORIGAMI as a composable motion primitive for multi-DoF soft robotic architectures. This system-level composability is enabled by embedding module connections and relative-motion constraints directly within the system architecture. MONORIGAMI realizes this integration by incorporating actuation chambers, load-bearing facets, compliant creases, passive joints, structural constraints, and fluidic routing within a single-material, single-print structure. The resulting architecture reduces the separate joints, transmissions, mechanical interfaces, and alignment steps required to combine multiple actuators. This monolithic integration therefore mitigates the actuator–joint integration complexity that can accumulate with increasing DoFs and allows locally programmed module motions to be composed directly into multi-DoF system-level behavior. Because the intended motion, stiffness distribution, and deformation constraints are encoded directly in the manufactured geometry, system configurations can also be tailored to target applications while preserving the locally programmed behavior of the individual modules.

The broader applicability of this design framework is reflected in three complementary robotic systems. FingerPrint's serial–parallel architecture spatially combines the directional motions of the actuator pairs, generating roll, pitch, yaw, and z-axis translation without separate rotational joints or transmission mechanisms. By collectively exploiting the compliance of multiple modules, the configuration also produces an additional translational DoF, enabling multi-axis cutaneous feedback within a compact and lightweight wearable form factor. In the delta-inspired joystick, a parallel arrangement of three modules produces directionally resolved kinesthetic feedback for teleoperation. In the modular gripper, serial combinations of modules with different geometric parameters generate distinct finger trajectories. These systems demonstrate that a common MONORIGAMI actuation principle can be extended into distinct system-level behaviors through module orientation, serial and parallel connection topologies, and heterogeneous geometric configurations.

Despite these advantages, several limitations remain to be addressed in future development. First, although monolithic fabrication reduces actuator-level assembly, the current process still requires post-processing, including residual resin removal from enclosed chambers, drainage-port sealing, post-curing, and support removal. These procedures may become increasingly difficult as the structures are miniaturized or as the number and complexity of internal channels increase; support-free geometries or 2D-to-3D folding strategies inspired by conventional origami robotics could reduce this processing burden. Second, single-material fabrication limits the achievable stiffness contrast between facets and creases, particularly for larger structures or applications requiring greater load-bearing capability. Multi-material 3D printing, alternative geometric stiffening strategies, or advanced printable materials could improve local stiffness programming and load-bearing performance without substantially increasing structural volume. Finally, sensing is not directly integrated into the current MONORIGAMI actuators, and most demonstrations rely on open-loop actuation. Embedding soft strain, pressure, or deformation sensors could enable closed-loop control that compensates for material nonlinearity, viscoelasticity, fabrication variation, and external disturbances. Therefore, future work will investigate actuator arrangements, sensing integration, multi-DoF configurations, and printable materials to improve load-bearing performance and control fidelity.

## MATERIALS AND METHODS

### MONORIGAMI design and fabrication

The MONORIGAMI actuator was designed using SolidWorks computer-aided design (CAD) software (Dassault Systèmes) and converted into a stereolithography (STL) file for 3D printing. PreForm software was used to generate the support structures required for the printing process. The actuator was fabricated using a Form 3 SLA printer (Formlabs), selected for its low cost, high resolution, and capability to produce air-tight structures without additional processing. Flexible 80A resin was employed because of its flexibility, low extensibility, and high repeatability.

The fabrication process consisted of five main steps (Fig. S10).

**1. Monolithic Fabrication:** The entire device was fabricated monolithically using the SLA 3D printer.

**2. Removing Residual Resin:** Residual resin trapped within the internal chambers was removed through the drainage ports using a vacuum compressor and syringe needle.

**3. Sealing the Drainage Ports:** The drainage ports were filled with uncured resin and partially cured using UV light to seal the openings.

**4. Curing the Entire Device:** The device was subsequently post-cured at 60 °C for 4 min to ensure complete polymerization.

**5. Removing Supports:** The support structures generated during the printing process were manually removed from the partially post-processed device.

## Actuator Characterization Test Setup

**Blocking torque measurement:** Blocking torque was measured using an ATI Nano 17 F/T sensor (ATI Industrial Automation), as shown in Fig. S4. The blocking position was controlled by a DC motor with an encoder (Resolution: 1024 pulses per revolution). The blocking torque ($\tau$) was estimated using the measured force in the x-axis ($F_x$) and calculated with the following equation:

$$\tau = r\frac{F_x}{\sin(\Delta\theta_f)} \quad (3)$$

where $r$ is joint center-contact point distance. Experiments were performed at 3° intervals from $\Delta\theta_f$ = 0° under pressures of −20, −40, −60, and −80 kPa, with $N_{rept}$ = 10 repetitions. To measure the steady-state blocking torque, a 1 Hz square wave form was used to apply the pressure.

**Bandwidth measurement:** Actuator motion was tracked using a color marker at the tip (Fig. S2), recorded at 240 frames per second (FPS) and analyzed with OpenCV (Python). The response magnitude was calculated as follows:

$$M_{dB} = 20\log_{10}\left(\frac{\Delta\theta_f}{\Delta\theta_{f,max}}\right) \quad (4)$$

Tests were conducted at 0.0625, 0.125, 0.25, 0.5, 1, and 2 Hz ($N_{rept}$ = 10 each).

**Hysteresis:** Hysteresis was measured using the same color-marker tracking method, with pressure varied in 5-kPa increments under a 1-Hz square-wave input.

**Load-bearing test:** Using the same tracking method, an actuator with $\theta_d$ = 80° and $t$ = 0.3 mm was tested with loads of 0, 20, 50, and 100 g under a 1-Hz square-wave pressure input.

## Virtual Reality Test Environment

**Version 1 (Fig. S8A).** The simulation was implemented in MATLAB on Microsoft Windows. IMU-derived roll, pitch, and yaw data were transmitted to the PC via serial communication and used to compute reaction forces and directions. The resulting control commands were sent through serial communication to an Arduino Teensy to actuate the solenoids.

Within the simulation environment, finger pose and contact force were calculated during interaction with virtual objects. The finger IMU was mapped one-to-one to simulated orientation, while the dorsal-hand IMU was scaled such that 1° corresponded to 10 mm of z-axis translation. Signals were generated only upon virtual contact and sent to a microcontroller, which actuated designated MONORIGAMI unit pairs when predefined thresholds were exceeded (−12 mm in z- axis translation, ±20° roll, ±15° pitch, ±10° yaw).

**Version 2 (Fig. S8B).** The virtual environment and all interaction forces are simulated using the CHAI3D haptics and simulation framework *(42)*. The user interacts with the environment via virtual avatars in the form of finger-shaped meshes representing the user's index finger and thumb. Upon contact, the FingerPrints are given an air pressure command proportional to the magnitude of calculated interaction forces.

Prior work focusing on user performance during two-finger manipulation task demonstrated this type of manipulation and control with traditionally servo-actuated devices *(43)*, as well as with other devices that are similarly pneumatically actuated as the FingerPrint *(44)*. The virtual environment can be customized by prescribing the mechanical properties such as mass, friction, and effective stiffness, or a lack thereof of the virtual objects presented to the user. For example, in the virtual environment presented in Fig. S3B, there is a combination of 1) dynamic virtual objects (the cube) where the user can interact with the object and receive interaction forces and therefore haptic feedback, and 2) purely visual objects such as the hoops and target area intended to guide the user in completing the prescribed task.

### Teleoperation Test Environment

The teleoperation system was established via TCP/IP communication between the follower robot (Panda, Franka Emika) and a PC. The follower was controlled using a position–velocity controller implemented in ROS 2 Humble on Ubuntu 22.04 (Fig. S9). On the task side, the follower's position, force, and torque data are sent back to the operation side, allowing the joystick to provide kinesthetic feedback to the operator based on this data (Fig. S9A). The joystick's position was measured at a frequency of 60 Hz using a 3D Hall-effect sensor (From SparkFun Electronics). On the operation side, the PC transmits the joystick's position to the follower via the network, and the follower is velocity-controlled based on this data (Fig. S9B). The interaction between the follower and the environment was measured using an Axia80 sensor (from ATI Industrial Automation) at a frequency of 1 kHz. For the virtual wall test, an AFT20-15D sensor (AIDIN ROBOTICS) was mounted on the joystick to measure kinesthetic feedback. In the environment contact test, the joystick position was tracked using a color marker method under the same conditions as the characterization experiments.

**Acknowledgments:**

**Funding:** This work was supported by the National Research Foundation of Korea (NRF) grant funded by the Korean government (MSIT) in 2026 (Grant No. RS-2025-02213804, Project Title: In-Space Servicing and Manufacturing Research Center)); the Swiss National Science Foundation (SNSF) Early Postdoc Mobility grant P2ELP2 195132; U.S. National Science Foundation grants 1830163 and 1812966; the U.S. National Science Foundation Graduate Research Fellowship Program; the Link Foundation Fellowship; the Black in Robotics Legacy Fellowship; and Sony Corporation of America.
**Author contributions:** J.J. and Z.Z. conceived the research, developed the MONORIGAMI concept and methodology, designed the robotic systems, analyzed and interpreted the results, prepared the figures, and wrote the original manuscript. J.E.P. developed and performed the simulations and contributed to the analysis and interpretation of the simulation results. M.K. assisted with experimental design, conducted experiments, and contributed to data analysis and validation. J.H.R. and A.M.O. supervised the research, provided technical and intellectual guidance, acquired funding, and reviewed and edited the manuscript. All authors discussed the results and approved the final manuscript.
**Competing interests:** The authors declare that they have no competing interests.
**Data and materials availability:** All data needed to evaluate the conclusions in the paper, including the CAD files for the robotic system and the raw experimental data, are available in the paper or the Supplementary Materials.